\documentclass[utf8]{FrontiersinHarvard} % for articles in journals using the Harvard Referencing Style (Author-Date), for Frontiers Reference Styles by Journal: https://zendesk.frontiersin.org/hc/en-us/articles/360017860337-Frontiers-Reference-Styles-by-Journal
\usepackage{url,hyperref,lineno,microtype}
\usepackage[onehalfspacing]{setspace}

\usepackage{cite}
\usepackage{amsmath,amssymb,amsfonts}
\usepackage{algorithmic}
\usepackage{graphicx}
\usepackage{textcomp}
\usepackage{booktabs}
\usepackage{svg}
\usepackage{comment}
\usepackage{adjustbox}
\usepackage{array}
\usepackage{balance}
\usepackage{multirow}
\usepackage{xspace}
\usepackage{graphicx}
\usepackage{color, soul}

\usepackage{booktabs}
\usepackage{siunitx}

\def\keyFont{\fontsize{8}{11}\helveticabold }
\def\firstAuthorLast{Zandigohar {et~al.}} %use et al only if is more than 1 author
\def\Authors{Mehrshad Zandigohar\,$^{1}$, Mo Han, Mohammadreza Sharif\,$^{1}$, Sezen Ya{\u{g}}mur G{\"{u}}nay\,$^{1}$, Mariusz P. Furmanek\,$^{1}$, Mathew Yarossi\,$^{1}$, Paolo Bonato\,$^{2}$, Cagdas Onal\,$^{3}$, Ta\c{s}k{\i}n Pad{\i}r\,$^{1}$, Deniz Erdo{\u{g}}mu{\c{s}}\,$^{1}$, Gunar Schirner\,$^{1,*}$}

\def\Address{$^{1}$ Northeastern University, Boston, MA, USA \\
$^{2}$Motion Analysis Lab, Spaulding Rehabilitation Hospital, Charlestown, MA, USA \\
$^{3}$Soft Robotics Lab, Worcester Polytechnic Institute, Worcester, MA, USA  }
\def\corrAuthor{Gunar Schirner}

\def\corrEmail{schirner@ece.neu.edu}

\begin{document}
\onecolumn
\firstpage{1}

\title {Multimodal Fusion of EMG and Vision for Human Grasp Intent Inference in Prosthetic Hand Control} 

\author[\firstAuthorLast ]{\Authors} %This field will be automatically populated
\address{} %This field will be automatically populated
\correspondance{} %This field will be automatically populated

\extraAuth{}% If there are more than 1 corresponding author, comment this line and uncomment the next one.
%\extraAuth{corresponding Author2 \\ Laboratory X2, Institute X2, Department X2, Organization X2, Street X2, City X2 , State XX2 (only USA, Canada and Australia), Zip Code2, X2 Country X2, email2@uni2.edu}

\maketitle

% \begin{abstract}

% %%% Leave the Abstract empty if your article does not require one, please see the Summary Table for full details.
% \section{}
% For full guidelines regarding your manuscript please refer to \href{https://www.frontiersin.org/guidelines/author-guidelines}{Author Guidelines}.

% As a primary goal, the abstract should render the general significance and conceptual advance of the work clearly accessible to a broad readership. References should not be cited in the abstract. Leave the Abstract empty if your article does not require one, please see the Article Types on every Frontiers journal page for full details

% \tiny
%  \keyFont{ \section{Keywords:} keyword, keyword, keyword, keyword, keyword, keyword, keyword, keyword} %All article types: you may provide up to 8 keywords; at least 5 are mandatory.
% \end{abstract}

\begin{abstract}
Objective: For transradial amputees, robotic prosthetic hands promise to regain the capability to perform daily living activities. Current control methods based on physiological signals such as electromyography (EMG) are prone to yielding poor inference outcomes due to motion artifacts, muscle fatigue, and many more. Vision sensors are a major source of information about the environment state and can play a vital role in inferring feasible and intended gestures. However, visual evidence is also susceptible to its own artifacts, most often due to object occlusion, lighting changes, etc. Multimodal evidence fusion using physiological and vision sensor measurements is a natural approach due to the complementary strengths of these modalities. Methods: In this paper, we present a Bayesian evidence fusion framework for grasp intent inference using eye-view video, eye-gaze, and EMG from the forearm processed by neural network models. We analyze individual and fused performance as a function of time as the hand approaches the object to grasp it. For this purpose, we have also developed novel data processing and augmentation techniques to train neural network components. Results: Our results indicate that, on average, fusion improves the instantaneous upcoming grasp type classification accuracy while in the reaching phase by 13.66\% and 14.8\%, relative to EMG  (81.64\% non-fused) and visual evidence  (80.5\% non-fused) individually, resulting in an overall fusion accuracy of 95.3\%. Conclusion: Our experimental data analyses demonstrate that EMG and visual evidence show complementary strengths, and as a consequence, fusion of multimodal evidence can outperform each individual evidence modality at any given time.

\tiny
 \keyFont{ \section{Keywords:} Dataset, EMG, Grasp Detection, Neural Networks, Robotic Prosthetic Hand} %All article types: you may provide up to 8 keywords; at least 5 are mandatory.

\end{abstract}

\section{Introduction}
In 2005, an estimated of 1.6 million people (1 out of 190 individuals) in the US were living with the loss of a limb \citep{ushand}. This number is expected to double by the year 2050. The most common prosthesis in upper extremity amputees is cosmetic hand type with the prevalence of 80.2\%  \citep{upperamput}. As limb loss usually occur in the working ages, dissatisfaction in the effectiveness of the prescribed prosthesis, will often impose difficulties in an amputee's personal and professional life. Therefore, providing a functional prosthesis is critical to address this issue and can improve quality of life for amputees.

In recent years there have been numerous efforts to leverage rapid advancements in machine learning \citep{pouyanfar2018survey, sunderhauf2018limits, mech} for intuitive control of a powered prosthesis.  Much of this effort is focused on inference of the human's intent using physiological signals from the amputee including electromyography (EMG) and to lesser extent electroencephalography (EEG)\citep{eeg1, emg1, emg2}. Myoelectric control, using EMG, of an upper limb prosthesis has been the subject of intense study and there are now several devices commercially available (for review see \citep{guo2023towards}). Despite the advances in myoelectric control and commercial availability of robotic prosthetic hands, current control methods generally lack robustness which reduces their effectiveness in amputees' activities of daily life \citep{kyranou2018causes}. Reliance solely on physiological signals from the amputee (such as EMG),  has many drawbacks that will adversely impact the performance of the prosthetic hand. These include artifacts caused by electrode shifting, changes of skin electrode impedance over time, muscle fatigue, cross-talk between electrodes, stump posture change, and the need for frequent calibration \citep{emgbad1,emgbad2}. Therefore, there is a need for additional sources of information to provide more robust control of the robotic hand.

\begin{figure}[t]
  \centering
  \includegraphics[width=0.48\textwidth]{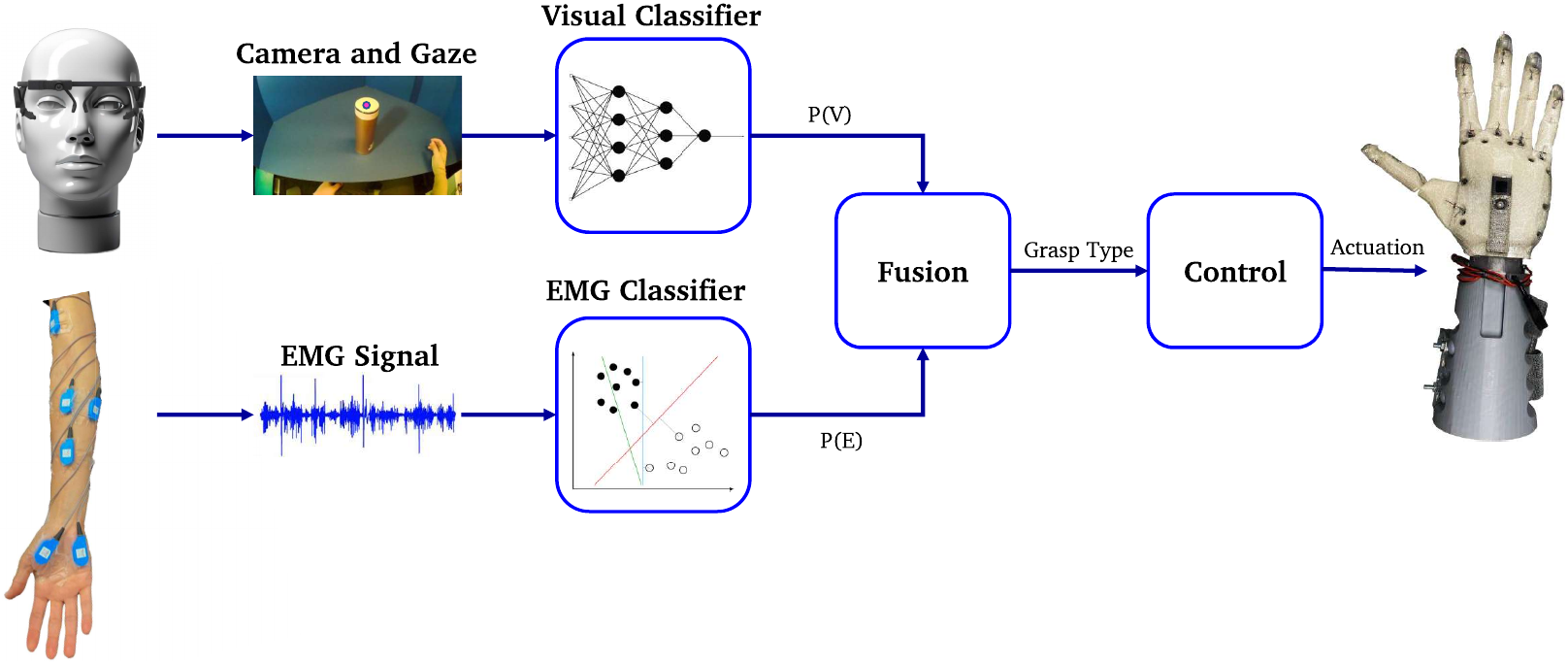}
  \caption{Proposed System Overview (eye-tracker from \citep{eyetracker}).}
  \vspace{-0.2cm}
\label{fig:sys_overview}
\end{figure}

A second major source of information for state-of-the-art control of powered prosthesis are RGB cameras bundled with a control methods based on pattern recognition or deep neural networks. These methods generally use image information to infer the reaching trajectory, time of triggering the grasp or most importantly, the grasp type used to inform finger preshaping movements of the robotic hand \citep{relatedVis,robotgrasp,han2020hands, shi2020computer, park2022grasping}. Although great progress has been made in the use of convolutional neural networks to classify grasp type, most do not incorporate information about the gaze of the user. The lack of such data renders these methods incapable of providing the correct grasp type in clutter fields of view containing two or more objects. When multiple objects are present in the field of view, then the gaze modality is absolutely necessary to identify the intended object. Critically, similar to grasp classification using the physiological data (such as EMG evidence), solely relying on visual data with eye-gaze is still susceptible to artifacts such as object occlusion and lighting changes that limit the robustness of the system.

The work in \citep{cognolato2022improving} is similar in that it utilizes gaze information and EMG modalities. While the work in \citep{cognolato2022improving} only provides rest and non-rest phases, our work goes beyond that as it not only focuses on the grasp intent inference but paves the path for an actual robotic implementation by more closely considering the phases of interaction with the manipulated object including an estimation of when to start and stop interacting with the object as given the ability to detect all the 4 phases involved in handling objects i.e., reach, grasp, return and rest. This  is consistently reflected throughout our work, manifesting in the aspects of protocol design, selection of inference data, and the methodology of inference and fusion and is most noticeable in our discussion of accuracy over time. Lastly, our experimental protocols are designed based on a more dynamic protocol where the objects are moved around the table. On the other hand, the aforementioned work’s protocol is less sophisticated where the subject returns the object to the same position as the object was placed initially. As a result, our has been evaluated on a broader range of real-world scenarios, providing a more comprehensive and realistic foundation for research and analysis.

To increase the robustness of grasp classification, we propose fusing the evidence from amputee's physiological signals with the physical features evident in the visual data from the camera while in reaching phase. As presented in \autoref{fig:sys_overview}, the proposed system design consists of a neural visual classifier to detect and provide probabilities of grasp gestures given world imagery and eye-gaze from the eye tracking device; an EMG classifier predicting the EMG evidence from amputee's forearm including both phase detection and grasp classification; and a Bayesian evidence fusion framework to fuse the two. The selected gesture is then utilized by the robotic controller to actuate the fingers.

In addition, we approach the problem with a more dynamic protocol. The work in \citep{cognolato2022improving} is based on grasping an object and returning it to the same location. On the contrary, our approach involves the subjects carrying around the object in the environment, making it more sophisticated for both visual and EMG modalities resulting in more challenging data. As a result, our dataset encapsulates a broader range of real-world scenarios, providing a more comprehensive and realistic foundation for research and analysis.

Finally, we propose that employing simpler classification methods like Yolo, tailored for grasp type recognition, is not only effective but also offers faster inference times due to reduced computational complexity compared to Mask-RCNN and LSTM, thereby mitigating the risk of overfitting.

Our experimental results show that fusion can outperform each of the individual EMG and visual classification methods at any given time. Specifically, fusion improves the average grasp classification accuracy during reaching by $13.66\%$ (81.64\% non-fused), and $14.8\%$ (80.5\% non-fused) for EMG and visual classification respectively with a total accuracy of $95.3\%$. Moreover, such utilization of fusion allows the robotic hand controller to deduce the correct gesture in a more timely fashion, hence, additional time is left for the actuation of the robotic hand. All classification methods were tested on our custom dataset with synchronized EMG and imagery data. The main contributions of this work are:
\begin{itemize}
    \item \textbf{Synchronized grasp dataset:} We collected a multimodal dataset for prosthetic control consisting of imagery, gaze and dynamic EMG data, from 5 subjects using state-of-the-art sensors, all synchronized in time.
    \item \textbf{Grasp segmentation and classification of dynamic EMG:} 
    We segmented the non-static EMG data into multiple dynamic motion sequences with an unsupervised method, and implemented gesture classification based on the dynamic EMG.
    \item \textbf{Visual grasp detection:} 
    We built a CNN classifier capable of detecting gestures in visual data, and background generalization using copy-paste augmentation.
    \item \textbf{Robust grasp detection:} 
    We implemented the multimodal fusion of EMG and imagery evidence classifications, resulting in improved robustness and accuracy at all times. 
\end{itemize}

The rest of this paper continues as follows: \autoref{sec:system} provides details on system setup and data collection protocols. After that, \autoref{sec:emg} provides an in-depth study of EMG phase segmentation and gesture classification. Then, \autoref{sec:vision} discusses visual detection and generalization methods. Afterwards, \autoref{sec:fusion} elaborates on our fusion formulation. In \autoref{sec:results} we describe the metrics used and present the results for EMG, visual and fusion systems. Moreover, we discuss related works, limitations and advancements in \autoref{sec:discussion}. Finally, we conclude our work in the \autoref{sec:conclusion}.

\section{System Setup}
\label{sec:system}

This section provides the technical details required to replicate the data acquisition system and results. The details are provided in four subsections: 1) system overview, 2) sensor configurations, 3) experiment protocol, and 4) data collection.

\subsection{System Overview}

Our data acquisition system entails collecting, synchronizing, and storing information from subject's eye-gaze, surface EMG, and an outward facing world camera fixed to the eye-glass worn eye-tracker. Robot Operating System (ROS) enabled the communication framework. ROS Bag file format was used to store the data on disc.

In our exploration of camera configurations, we initially implemented a palm-mounted camera in our previous works \citep{han2020hands, zandigohar2019towards}, but practical constraints prompted a transition away from this setup. A significant challenge surfaced as the object within the camera's field of view became visible only briefly and relatively late in the grasping process, allowing insufficient time for preparatory actions. To optimize the efficacy of the palm camera, we found it necessary to prescribe an artificial trajectory for the hand. Regrettably, this imposed trajectory compromised the natural and intuitive elements of the interaction. Consequently, we pivoted towards a head-mounted camera in our current work with the following advantages: (1) a head-mounted camera provides a broader and more comprehensive field of view and therefore enhances the prosthetic hand's ability to interpret the context of the grasp. Understanding the spatial context is crucial for executing precise and contextually relevant grasps. For example, if the prosthetic hand needs to grasp an object on a shelf, the head-mounted camera can capture not only the target object but also the spatial relationship between the hand and the shelf. This information aids in determining the appropriate hand posture required for a successful grasp in that specific spatial context. (2) A camera on the head offers more stability and consistency in capturing visual input. The head, being a relatively stable platform, ensures that the camera maintains a consistent orientation, distance and perspective which is crucial for accurate and reliable grasp detection. (3) Placing the camera on the head aligns more closely with the user's natural way of perceiving and interacting with objects. People tend to look at objects of interest before reaching for them, and a head-mounted camera can mimic this natural behavior, facilitating more intuitive grasp detection. A palm-mounted camera on the other hand is heavily dependent on the reach to grasp trajectory and is adversely affected by variations in hand orientation or distance to the object during grasping. For example, when reaching to put an upside down glass in the dishwasher an individual would like reach in a supinated position with the palm facing the floor. In this case a palm mounted camera would not have view of the glass until just prior to grasp.

It is important to recognize that while a palm-mounted camera offers distinct benefits, such as delivering detailed insights into the reaching conditions, including wrist rotation, which can be essential for achieving greater autonomy in prosthetic control, this aspect falls outside the scope of our current study. Our research is specifically concentrated on detecting the type of grasp, for which the head-mounted camera is more aptly suited. This focus on grasp type detection aligns the head-mounted camera's broader field of view and user-aligned perspective with our study's objectives, making it a more appropriate choice in this context.

A mobile binocular eye-tracker (Pupil Core headset, PupilLab, Germany) with eye facing infrared cameras for gaze tracking and a world facing RGB camera was used. The gaze accuracy and precision were 0.60° and 2\%, respectively. The gaze detection latency was at least $11.5$ ms according to the manufacturer. The world camera of the eye-tracker recorded the work-space at 60Hz@720p and FOV of 99°x53°. The gaze and world camera data were sent to ROS in real-time using ZeroMQ \citep{ZeroMQ}.

Muscle activity was recorded from subjects' right forearm through 12 Z03 EMG pre-amplifiers with integrated ground reference (Motion Lab Systems, Baton Rouge, LA, USA). The pre-amplifiers provided x300 gain and protection against electrostatic discharge (ESD) and radio-frequency interference (RFI). The signals were passed to two B\&L 6-channel EMG electrode interfaces (BL-EMG-6). Then, an ADLINK USB 1902 DAQ was used to digitize EMG data, which was then stored along with other signals in the ROS Bag file. The ADLINK DAQ used a double-buffer mechanism to convert analog signals. Each buffer was published to ROS when full. The system components are depicted in \autoref{fig:system_a}.

\subsection{Sensor Configurations}

\setcounter{subfigure}{0}
\begin{subfigure}
\setcounter{subfigure}{0}
    \centering
    \begin{minipage}[b]{0.45\textwidth}
       \includegraphics[width=0.5\textwidth]{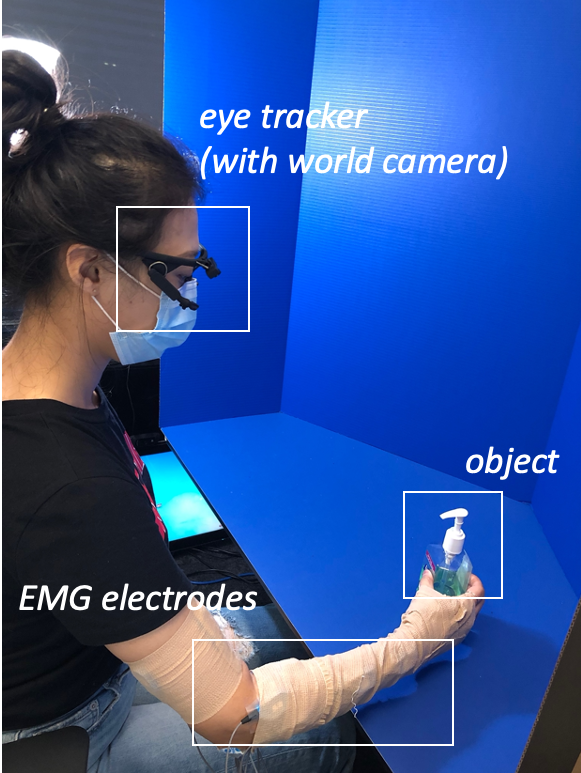}
       \caption{System components}
       \label{fig:system_a}
    \end{minipage}%
\setcounter{subfigure}{1}
\begin{minipage}[b]{0.45\textwidth}
   \includegraphics[width=\textwidth]{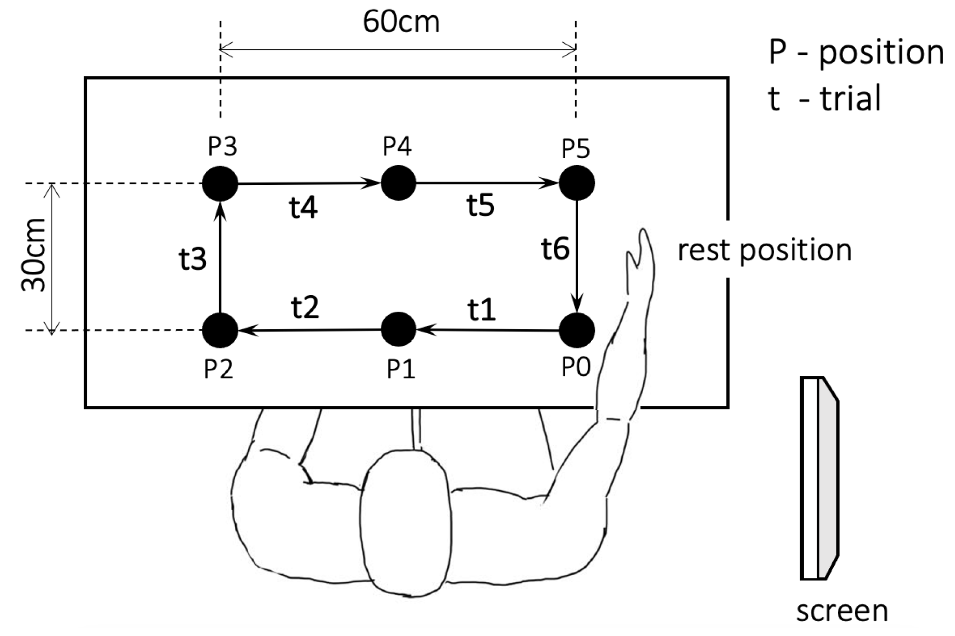}
   \caption{Data collection setup}
   \label{fig:system_b}
\end{minipage}

\setcounter{subfigure}{-1}
\caption{System components and data collection setup. Only the clockwise session is demonstrated in (b).}
\label{fig:system}
\end{subfigure}

\subsubsection{Eye-tracker Configuration}

The orientation of the eye and world cameras were adjusted at the beginning of the experiment for each subject and remained fixed during the whole experiment. A single marker calibration method leveraging the vestibulo-ocular reflex (VOR) was used to calibrate the gaze tracker. While gazing at the marker lying on the table, the subject moved their head slowly to cover the whole field of view. 

\subsubsection{EMG Sensor Configuration}

Surface EMG was recorded at $f=1562.5$ Hz in $C=12$ muscles of the arm, forearm and hand in order to capture dynamic hand gesture information and arm movement: First Dorsal Interosseous (FDI), Abductor Pollicis Brevis (APB), Flexor Digiti Minimi (FDM), Extensor Indicis (EI), Extensor Digitorum Communis (EDC), Flexor Digitorum Superficialis (FDS), Brachioradialis (BRD), Extensor Carpi Radialis (ECR), Extensor Carpi Ulnaris (ECU), Flexor Carpi Ulnaris (FCU), Biceps Brachii-Long Head (BIC), and Triceps Brachii-Lateral Head (TRI). The muscle locations were found by palpation during voluntary arm movements. After skin preparation, the surface electrodes were fixed to the skin overlying each muscle using tape. 

\begin{figure}[t!]
  \centering
  \includegraphics[width=0.48\textwidth]{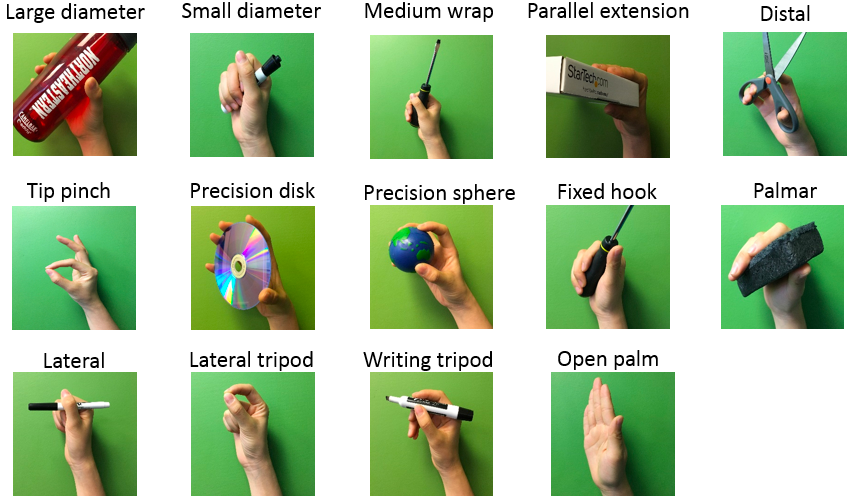}
  \caption{Selected $14$ gestures for the classification problem.}
  \vspace{-0.2cm}
\label{fig_14grasps}
\end{figure}

\subsection{Experiment Protocol}

Enabling online human-robot interaction via hand and arm motion proves challenging due to the intricate structure and high degree of freedom (DOFs) present in the human body. Specifically, the human hand is complex with its 21 DOFs and 29 controlling muscles. As such, recognizing real-time human grasp intentions through the identification of intricate and high-DOF hand motions poses a significant challenge. 

Historically, research has mainly focused on steady-state classification involving a limited number of grasp patterns, which proves inadequate when addressing the nuanced changes in muscle activity that occur in real-world scenarios. For instance, conventional protocols usually instruct participants to maintain a specific gesture for a period of time, capturing data during this static process. However, in practical applications, muscular activity and hand configuration shift between static and dynamic positions, changing in unison with arm movements. 

To improve the detection of dynamic movements, we have incorporated greater variability into our experimental protocol. This involved gathering and employing data from an expanded repertoire of gestures, and integrating multiple dynamic actions to represent the transitions between various grasp intentions based on natural human movement sequences. Specifically, our experiment synchronizes hand movements with dynamic arm motions. We ask participants to naturally and continuously grasp different target objects from various orientations and positions without a pause. As the hand moves towards a target to grasp it, the finger and wrist configurations dynamically adapt based on the object's shape and distance. By including all these movement phases and changes in our experiment, we collect a broad range of motion data. This approach allows us to leverage the continuity of hand formation changes to enhance data variability and mimic real-life scenario.

The experiment included moving objects among cells of an imaginary 3$\times$2 grid with pre-defined gestures. The experimental setup is shown in \autoref{fig:system} and the pre-defined gestures are shown in \autoref{fig_14grasps}. The experiment was comprised of two sessions. In the first session, subjects moved objects in a clockwise manner, and in the second session objects were moved in a counter-clockwise manner (only the clockwise session is demonstrated in \autoref{fig:system}). Having two sessions contributed to the diversity of the dataset for the EMG signal and the camera image patterns.

In each session, 54 objects were placed on a table one by one within the reach of the subjects so they could locate it on the proper spot before the experiment began. Then, for each object, an image was shown on the monitor on the right side of the subject instructing them how to grasp the object. Then, the moving experiment started. Audio cues, i.e. short beep sounds, were used to trigger each move. 

The subject performed $6$ trials for each object, where each trial was executed along its corresponding predefined path, as shown in \autoref{fig:system_b}. During the first trial t1, the object was moved from the initial position P0 to the position P1, followed with another five trials to move the object clockwise until it was returned to the initial position P0, leading to $6$ trials in total. After moving all objects in the clockwise order, the second session of counterclockwise started after a 15-minute break. The break meant to help the subject refresh and to maintain focus on the task in the second session. The interval between consecutive audio cues was set to 4 seconds for all objects. The entire experiment for the 6 trials per object, i.e. moving one object around the rectangle, took about 37 seconds including the instructions. In our experiments, each trial t, constitutes all four distinct phases. Participants begin each trial in a resting position and then proceed to reach for an object. Following the reach, they execute a grasp and transport the object to its next predetermined location. Upon successfully placing the object at its designated spot, participants release it and return their hand to the initial resting position, thus marking the completion of the trial and the re-entry into the resting phase.

\subsection{Data Collection}
\label{subsection:data_collection}

% THE FOLLOWING MUST BE COMMENTED:
% All experiments were conducted in accordance with the declaration of Helsinki, and the study design was approved by the Institutional Review Board of Northeastern University. 

Our decision to initiate our experiments with healthy subjects is informed by three primary factors. Firstly, identifying disabled or amputated participants is more challenging due to a smaller available pool and safety considerations. Additional time, effort, and administrative procedures are typically required to facilitate experiments involving these individuals. Secondly, before we consider involving disabled participants, it is important that we establish preliminary evidence of the effectiveness and feasibility of our method. By observing strong patterns among healthy subjects, we can utilize these findings as foundational data to extend similar experiments to disabled participants. Lastly and most importantly, the level of amputation can differ significantly among disabled individuals, complicating the process of collecting consistent EMG data channels. One potential solution to this problem is to use the model developed from the healthy subjects' data as a starting point. 

Experimental data were collected from 5 healthy subjects (4 male, 1 female; mean age: 26.7 ± 3.5 years) following institutionally approved informed consent. All subjects were right-handed and only the dominant hand was used for the data collection. None of the subjects had any known motor or psychological disorders. 

Feix et al.~\citep{graspTypes} proposed that human grasp taxonomy consists of 33 classes if only the static and stable gestures are taken into account. The human hand has at least $27$ degrees of freedom (DoF) to achieve such a wide range of gestures; however, most existing prosthetic hands do not have this many DoF~\citep{resnik2014deka}. Therefore, in our work, the experimental protocol was focused on those $14$ representative gestures involving commonly used gestures and wrist motions~\citep{graspTypes}. As shown in \autoref{fig_14grasps}, the $14$ classes were: large diameter, small diameter, medium wrap, parallel extension, distal, tip pinch, precision disk, precision sphere, fixed hook, palmar, lateral, lateral tripod, writing tripod, and open palm/rest. In our classification problem, we mapped the $14$ gestures with $14$ gesture labels $l \in {\{0,1,...,13\}}$, where $l=0$ was defined as open-palm/rest gesture and $l \in {\{1,...,13\}}$ were accordingly identified as the other $13$ gestures in the order listed in \autoref{fig_14grasps}.

\section{Classification of EMG Evidence}
\label{sec:emg}

% 0. data preprocessing

% 1. dynamic EMG classification over gesture
% 	a. training model using EMG from the reaching phase
% 	b. training model using EMG from the grasping phase
% 	c. training model using EMG from the reaching+grasping phase
	
% 2. Grasp trigger detection 

% 3. unsupervised segmentation of hand/arm movement

% 4. performance measurements over time

% * alignment of trials

% Future: which models we used to explore?

\begin{figure}[t]
  \centering
  \includegraphics[width=0.6\textwidth]{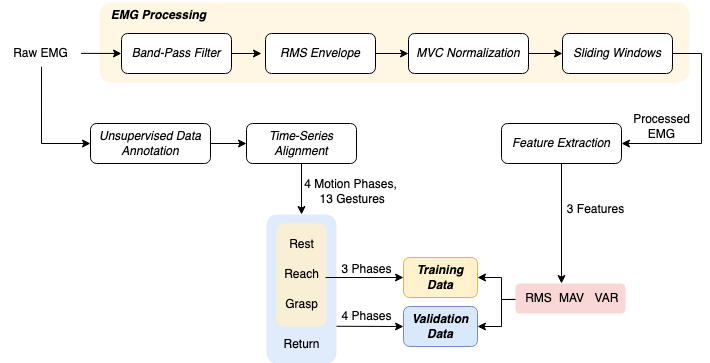}
  \caption{Schematic of the EMG Signal Processing and Data Annotation Workflow. This figure outlines the sequence of processing steps applied to EMG data, starting with band-pass filtering, followed by RMS envelope computation, MVC normalization, and the application of sliding window techniques. The featured extracted at this stage is used in gesture classification. In phase classifier, unsupervised data annotation and alignment provides phases of object manipulation – rest, reach, grasp, and return outlined in the lower section which constitute the labeled activities in the dataset for machine learning model training and validation, with features such as RMS, MAV, and VAR extracted for analysis.}
  \vspace{-0.2cm}
\label{fig_emg_overview}
\end{figure}

Extracting user hand/arm motion instructions from EMG signals has been widely utilized for human–robot interactions. A major challenge of online interaction with robots is the reliability of EMG recognition from real-time data. In this section, we introduce our method for the EMG control of the robotic hand. We propose a framework for classifying our collected EMG signals generated from continuous grasp movements with variations in dynamic arm/hand postures as outlined in \autoref{fig_emg_overview}. We first utilized an unsupervised segmentation method to segment the EMG data into multiple motion states, and then constructed a classifier based on those dynamic EMG data.

\subsection{EMG Data Pre-Processing}
\label{sub:emg_data_pre_processing}

As outlined in \autoref{fig_emg_overview}, first the raw EMG data were filtered with a band-pass Butterworth filter of $40$-$500$ Hz, where the high pass serves to remove motion artifact and the low pass is used for anti-aliasing and removal of any high frequency noise outside of the normal EMG range. Afterwards the root-mean-square envelopes~\citep{hogan1980rmsEnvelope} of the EMG signal were constructed using a sliding window of length $150$ samples (96ms). A maximum voluntary contraction (MVC) test was manually performed for each muscle at the beginning of the recordings. During the test, the subjects were instructed to perform isometric contractions constantly for each muscle \citep{kendall2005muscles}. Finally, the resulting EMG envelopes were normalized to the maximum window value of MVC data, which were processed the same as the task data. Finally, the filtered, RMS-ed and normalized EMG signals were further divided into sliding windows of $T=320$ ms, with an overlap of $32$ ms between two consecutive windows. Both feature extraction and classification were conducted based on each sliding window.

% The processed EMG signals were further divided into sliding windows of $T=320$ ms, with a delay of $32$ ms between two consecutive windows, where the selected lengths of window size and overlapping were shown to perform a better trade-off between decision delay and accuracy than other window configurations. Both classifications of movement phase and grasp type were conducted based on each window.

\subsection{EMG Feature Extraction}
% Features of EMG in the time domain can be extracted based on the raw EMG time series in real-time without any transformation, and require lower computational complexity compared with other features~\citep{phinyomark2012EMGfeature}. 

Following the pre-processing stage, wherein the raw EMG signal is subjected to band-pass filtering, RMS enveloping, MVC normalization, and segmentation into windows, we can proceed to extract features from this refined data.

Three time domain features were adopted in this work, including root mean square (RMS), mean absolute value (MAV), and variance of EMG (VAR)~\citep{phinyomark2012EMGfeature}. The RMS feature represents the square root of the average power of the EMG signal for a given period of time, which models the EMG amplitude as a Gaussian distribution. MAV feature is an average of absolute value of the EMG signal amplitude, which indicates the area under the EMG signal once it has been rectified~\citep{emg1}. VAR feature is defined as the variance of EMG, which is calculated as an average of square values of the deviation of the signal from the mean. In choosing our festure set, our experiments revealed that the selected set of features optimally suits our dataset, demonstrating enhanced generalization capabilities for unknown subjects, particularly in inter-subject and left-out validation. While these features may appear to encapsulate similar aspects of muscle activity, they each offer unique insights under different conditions which can be invaluable for robust classification in diverse scenarios. For instance, RMS provides a measure of the power of the signal, effectively capturing the overall muscle activity and is particularly sensitive to changes in force. MAV offers a quick and efficient representation of signal amplitude, useful for real-time applications and less sensitive to variations in signal strength compared to RMS. VAR reveals the variability in the signal, an important indicator of muscle fatigue and changes in muscle fiber recruitment patterns.

The input for the feature extraction is the pre-processed EMG window $X \in \mathbb{R}^{C \times T}$, where $C=12$ is the channel number of EMG from all muscles and $T=320$ ms is the window length with a sampling rate of $f=1562.5$ Hz. For each input EMG window $X \in \mathbb{R}^{C \times T}$, we extracted all the three mentioned time-domain features leading to an output feature vector of $Z \in \mathbb{R}^{3C \times 1}$.

\subsection{Data Annotation}
\label{sec: Data Annotation}

\begin{figure}[t]
  \centering
  \includegraphics[width=0.39\textwidth]{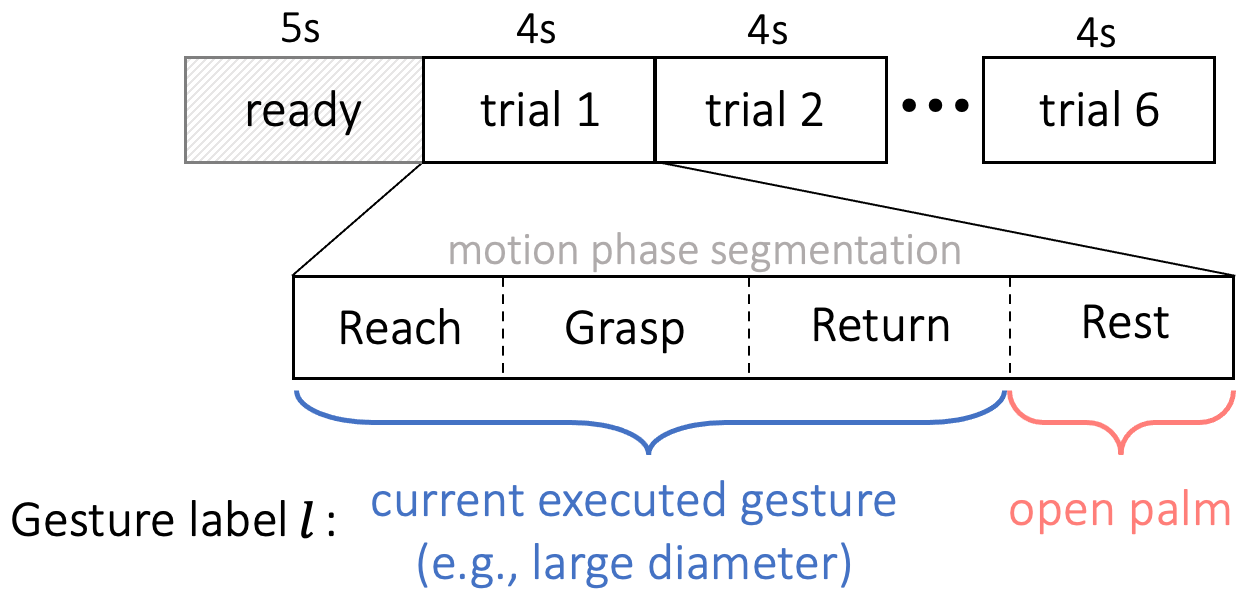}
%   \caption{Experiment timeline and the EMG segmentation and annotation.}
  \caption{Experiment timeline. The subject was given $5$ seconds to read the shown gesture before the first trial. Each trial lasted for $4$ seconds, repeated for $6$ trials without interruption. All EMG trials were segmented unsupervisedly into four sequences of reaching, grasping, returning and resting. The first three motion phases were labeled as gesture $l \in {\{1,...,13\}}$ corresponding to the target object, and the resting phase was tagged by the open-palm label $l=0$.}
  \vspace{-0.2cm}
\label{fig_timeline}
\end{figure}

In order to approach the gesture classification in a continuous manner, each EMG trial was assumed to include $4$ different movement sequences, i.e. reaching, grasping, returning and resting. The proposed motion sequences are naturally and commonly performed actions by human during the reach-to-grasp movements, giving greater probability to intent transitions that are likely to follow one another, such as a “grasp” action is always following a “reach” movement and followed by a "return" action. In our method, as shown in \autoref{fig_timeline}, we first segmented each EMG trial unsupervisedly into $4$ sequences, and then labeled them separately with gesture label $l$ according to the specific motion. During each trial, the dynamic grasp movements were performed naturally by the subject without limitation on the timing of each motion phase, so the length of each phase is not necessarily equal.

\subsubsection{Unsupervised EMG Segmentation of Dynamic Motion}
The EMG trial from dynamic grasp movement was segmented using an unsupervised method of Greedy Gaussian Segmentation (GGS)~\citep{hallac2019GGS}, based on the assumption that EMG signal under a specific stationary status can be well modeled as a zero-mean random process which is Gaussian distributed~\citep{clancy1999EMGgaussian}. 
The segmentation of dynamic motion was conducted using unprocessed EMG signals, ensuring that the raw data was directly employed without the application of any preliminary filtering or normalization procedures. The GGS method was proposed to solve the problem of breaking multivariate time series into segments over which the data is well explained as independent samples from different Gaussian distributions corresponding to each segment. GGS assumes that, in each segment, the mean and covariance are constant and unrelated to the means and covariances in all other segments. The problem was formulated as a maximum likelihood problem, which was further reduced to a optimization task of searching over the optimal breakpoints leading to the overall maximum likelihood from all Gaussian segments. The approximate solution of the optimized segments was computed in a scalable and greedy way of dynamic programming, by adding one breakpoint in each iteration and then adjusting all the breakpoints to approximately maximize the objective.

% \begin{figure}[t]
%   \centering
%   \includegraphics[width=0.35\textwidth]{figures/emg/ggs.png}
%   \caption{An example of the unsupervised motion phase segmentation of dynamic EMG signal using the GGS algorithm.}
% %   \caption{An example of the unsupervised motion phase segmentation of dynamic EMG signal using the GGS algorithm, where the EMG signal includes $C=12$ channels and the red dashed lines represent the locally optimal segment boundaries. Each of the four segmented EMG sequences was modeled as an independent multivariate Gaussian distribution with different means and variances, corresponding to the four grasp motion phases of reaching, grasping, returning and resting respectively.}
%   \vspace{-0.2cm}
% \label{fig_ggs}
% \end{figure}

In order to formulate four segments corresponding to the four grasp motion phases (reaching, grasping, returning and resting), we assigned three breakpoints to each EMG trial. Practically each of the four dynamic phases may not be strictly steady-state, but we nevertheless encode such transitions from one intent to another based on the proposed motion sequences considering the inherent stochastic nature of EMG signals. We then utilized the GGS algorithm to locate the locally optimal segment boundaries given the specific number of segments. 
The obtained three optimal segment boundaries led to four EMG sequences, where each of the four sequences was modeled as an independent $12$-channel multivariate Gaussian distribution with different means and variances.
% An example of such Gaussian segmentation is shown in \autoref{fig_ggs}, where each of the four EMG sequences was modeled as an independent $12$-channel multivariate Gaussian distribution with different means and variances.

\subsubsection{Hand Gesture Annotation of EMG}
In order to classify gestures from dynamic EMG signals in a real-time manner, following the motion phase segmentation, the resulting EMG segments were further annotated by a group of gesture label $l \in {\{0,1,...,13\}}$, where $l=0$ was defined as open-palm/rest gesture and $l \in {\{1,...,13\}}$ were accordingly identified as the other $13$ gestures listed in \autoref{fig_14grasps}.

During the reach-to-grasp movement, the configuration of the fingers and wrist changes simultaneously and continuously with the arm’s motion according to the shape and distance of the target object~\citep{jeannerod1984graspTiming}. For example, humans tend to pre-shape their hands before they actually touch the target object during a grasp, and this formation of the limb before the grasp is in direct relation with the characteristics of the target object \citep{jeannerod1984graspTiming}. Therefore, to accomplish a smooth interpretation of the grasping gesture, as presented in \autoref{fig_timeline}, we annotated unsupervisedly segmented sequences of reaching, grasping and returning as the executed gesture $l \in {\{1,...,13\}}$ corresponding to the target object, and tagged the resting phase with the open-palm label $l=0$.

\subsection{Gesture Classification of Dynamic EMG}
We constructed a model for classifying the gesture $l \in {\{0,1,...,13\}}$ of dynamic EMG signals with corresponding data pairs of $\{(X_i,l)\} _{i=1}^{n}$, where $X_i \in \mathbb{R}^{C \times T}$ is the $i$th EMG window with channel number $C=12$ and window length $T=320$ ms of $f=1562.5$ Hz sampling rate, and $n$ is the total number of windows. For each EMG window $X_i \in \mathbb{R}^{C \times T}$, three time-domain features of RMS, MAV and VAR were extracted as $Z_i \in \mathbb{R}^{3C \times 1}$, leading to data pairs of $\{(Z_i,l)\} _{i=1}^{n}$, which were the final inputs to train the grasp-type classifier.

We utilized the extra-trees method~\citep{geurts2006extraTrees} as the classifier, which is an ensemble method that incorporates the averaging of various randomized decision trees on different sub-samples of the dataset to improve the model performance and robustness. The number of trees in the extra-trees forest was set to be $50$ in this work, and the minimum number of samples required to split an internal node was set as $2$. In this method, each individual tree contains 13 output nodes representing grasp classes. To provide the final vector of grasp type probabilities, the output for all trees are averaged yielding a single vector with 13 probabilities.

\section{Visual Gesture Classification }
\label{sec:vision}

As mentioned earlier, visual grasp detection does not rely on the amputee's physiological signals, and therefore can provide more robustness to the system than EMG alone. Despite this major advantage, there are known challenges to classification and detection from visual data using deep learning. More specifically, the classifier needs to be invariant to environment changes such as the lighting, background, camera rotation and noise. Moreover, in the case of grasp detection, the final decision should be invariant to the object's color. 

To face these challenges, we utilize training with data augmentation techniques. Using the aforementioned set of data, a state-of-the-art pretrained object detector is fine-tuned for the purpose of grasp detection. The grasp detector then provides bounding boxes of possible objects to be grasped with the probabilities of each gesture. The box closest to the user's gaze is then selected as the object of interest, and the corresponding probabilities will be redirected to the fusion module. The details for each step is provided below.

\subsection{Generalization of the Background}

\begin{figure*}[t]
  \centering
  \includegraphics[width=1\textwidth]{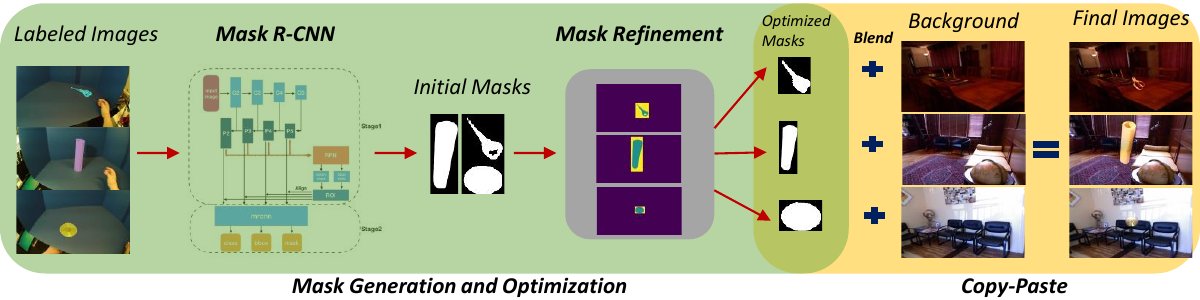}
  \caption{Overall overview of background generalization using visual mask generation and copy-paste augmentation.}
%   \vspace{-0.2cm}
\label{fig:overall_vision}
\end{figure*}

In order for a grasp detector to work properly in a variety of real-life scenarios and in different settings, it is crucial for it to be invariant to the background. Creating such a dataset is an arduous and somewhat impractical task since it requires access to many different locations and settings, and needs the participants and devices to be moved around. Therefore, in recent years, researchers \citep{egocentric, synthesizing} have used more practical solutions such as copy-paste augmentation \citep{copypaste} to tackle this issue. In copy-paste augmentation, using a mask which is usually obtained using a depth camera, the object of interest is copied from the background and pasted into a new background. This work aims to utilize copy-paste augmentation by relying only on the visual data. For this, a screen with a specific color such as blue or green is placed in the experimental environment as the background, and later chroma keying acquires a mask than can separate the foreground from background. Using this mask, the resulting foreground can be superimposed into image data from several places. The overall composition of the proposed copy-paste augmentation pipeline is demonstrated in \autoref{fig:overall_vision}.  

\subsubsection{Dataset for Background Images}
To have new backgrounds for this superimposition, we found the data from NYU Depth V2\citep{NYUV2} indoor scenes dataset to be very suitable. This dataset consists of images from 464 different, diverse and complex settings i.e., bedrooms, bathrooms, kitchens, home offices, libraries and many more that are captured from a wide range of commercial and residential buildings in three different US cities.

\subsubsection{Mask Generation}
During our experiments, we observed that most of the unsupervised computer vision methods which are usually based on color or intensity values fail to separate the foregrounds from the backgrounds correctly. With this intuition in mind, we found that instance segmentation is a more promising and robust method to obtain masks. Because of the simplicity of the task, even when using very few labeled data, retraining Mask R-CNN \citep{maskrcnn} can provide good enough masks to use in copy-paste augmentation. In our experiments, we labeled 12 images for each of the 54 objects totalling to 636 images. Each of the 12 images were constituted by selecting 2 random images from each of the 6 trials. To prevent over-fitting of the network to the very few data at at hand, they were heavily augmented using horizontal/vertical flipping, scaling, translation, rotation, blurring, sharpening, Gaussian noise and brightness and contrast changes. Moreover, we used ResNet-101 \citep{resnet} as the backbone structure.

\subsubsection{Refinement of Masks}
Although instance segmentation can provide correct bounding boxes and masks, it is crucial to have a very well defined mask when augmenting data with copy-paste augmentation. As seen in the original work \citep{maskrcnn}, despite the great success of Mask R-CNN in segmenting the objects, a closer look at the masks reveals that the masks do not match the objects' borders perfectly. This usually results in missing pixels in the destination image.

To further refine the masks to have more accurate borders, we propose to combine Mask R-CNN with GrabCut algorithm \citep{grabcut}. Each mask obtained by Mask R-CNN can be used as a \textit{definite foreground}, while anything outside the bounding box is considered \textit{definite background}. This leaves the pixels inside the bounding box that are not present in the initial mask as \textit{possible foreground}. 

\subsubsection{Blending}
\begin{figure}[t]
    \centering
    \begin{minipage}[b]{0.32\textwidth}
       \includegraphics[width=\textwidth]{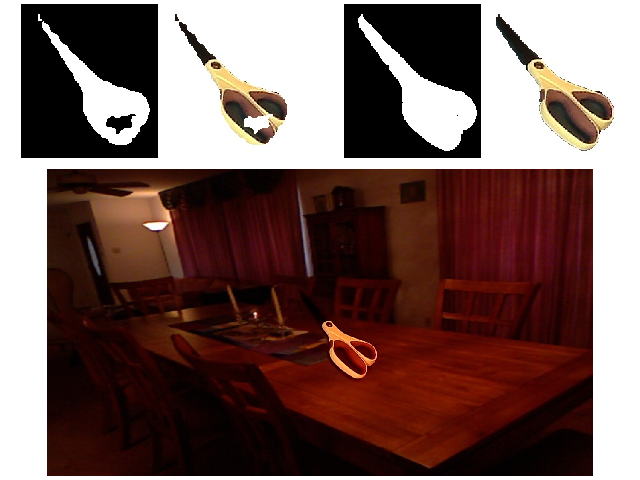}
    \end{minipage}%
\begin{minipage}[b]{0.32\textwidth}
   \includegraphics[width=\textwidth]{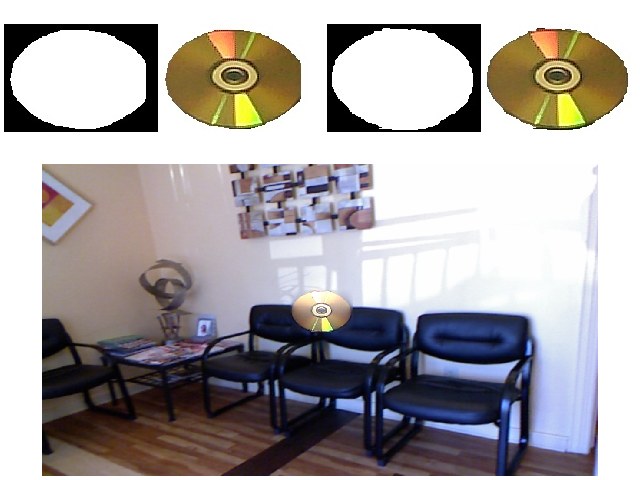}
\end{minipage}
    \centering
    \begin{minipage}[b]{0.32\textwidth}
       \includegraphics[width=\textwidth]{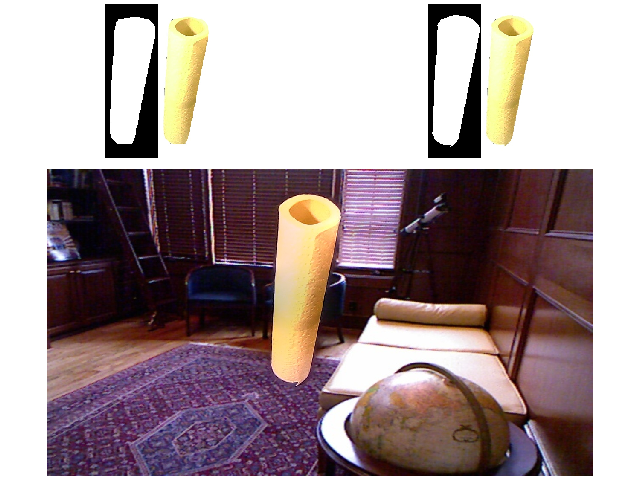}
    \end{minipage}%
\caption{Three examples of the final images after copy-paste augmentation. Images on the top demonstrate masks before and after refinement. As seen here, mask refinement can significantly improve mask borders and missing parts. Poisson blending can adjust the object being pasted w.r.t. the contrast and brightness of the destination image.}
\label{fig:blend_example}
\end{figure}

Due to contrast and lighting differences between the source and destination images, simply copying a foreground image does not result in a seamless final image. To have a seamless blend as the final step for copy-paste augmentation, we use Poisson blending \citep{poisson}. The refined masks from the previous step are slightly dilated to prevent null gradients. The resulting images are depicted in \autoref{fig:blend_example}.

Mask generation, refinement and blending provides a pipeline for generalizing the imagery data to enable classification of gestures in different environments. The rest of the augmentation techniques are mentioned in the next subsection accompanied by in-depth analysis of training the grasp detection and classification network.

\subsection{Gesture Detection and Classification}
In order to detect the suitable gesture from visual data and control the gesture of the robotic hand, the detector needs to find the box bounding the object and classify the possible gesture. We base our method on YoloV4 \citep{yolov4} that has shown promising results in the domain of object detection. YoloV4 is a fast operating speed object detector optimized for parallel computations in production developed in C++. The architecture of YoloV4 consists of: (i) backbone: CSPDarknet53, (ii) neck: SPP, PAN and (iii) head: YOLOv3. The similarity of MS-COCO dataset \citep{mscoco} which the network is trained on to our dataset makes YoloV4 a suitable source for transfer learning. As the COCO dataset is a general purpose dataset used in many classification, detection and segmentation networks specialized in objects purposed to be interacted with, we found its domain similar to our task at hand. This similarity makes it a suitable and opportune to exploit transfer learning. Out of the 54 object classes, 11 precisely match with classes from the COCO dataset. The remaining classes exhibit similarities either in appearance, such as pliers to scissors, or in context, for instance, "glass container (HANDS)" and "bowl (COCO)," both belonging to the kitchen appliances category. This nuanced categorization enhances the model's ability to recognize objects with contextual and visual resemblances. Lastly, the high throughput of the network when deployed on the GPU will result in real-time detection. Using this method, the object detector determines the bounding box corresponding to the person's gaze. The vector of probabilities for each grasp type is readily available in the last layer of the network. When used in isolation, the class with the highest probability is typically selected as the output of the object detector. However, in our case, we retain the probabilities for all classes and present the entire vector as our output. Therefore, we have a vector from the last layer of the YOLO neural network containing all 13 probabilities for each grasp type.

\section{Multimodal Fusion of EMG and Vision}
\label{sec:fusion}

Previous sections provided independent studies on classification of EMG and visual evidence, with the aim of providing generalized, realistic and accurate inference models for each source of information. Despite these efforts, there exists many contributors for each method to fail in the real world scenarios. To name but a few, EMG evidence would change drastically if any of the electrodes would shift, muscle is fatigued, skin electrode impedance is changed over time or a posture change. On the other hand, the visual information is similarly susceptible to its own artifacts, including object obstruction, lighting changes, etc. Fusion aims to improve robustness of the control method by exploiting multiple sources of information. 
% Not only a robust control can infer the intended grasp type more often, but also it allows the robotic hand to deduce the correct grasp type in a more timely fashion, hence, more time is left for the actuation of the grasp type. 
In this section, we first formulate the proposed fusion method and thereafter, validate and provide our results.

\begin{figure}[t]
  \centering
  \includegraphics[width=4cm]{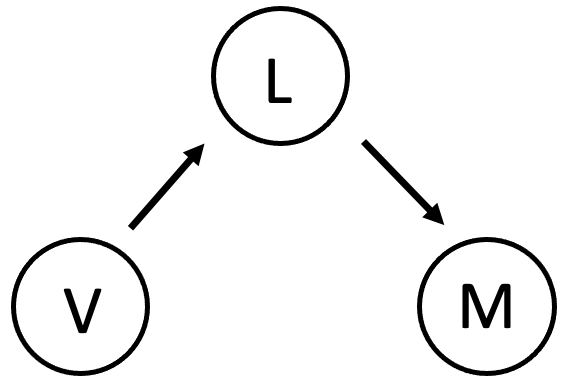}
\caption{The graphic model of the multimodal fusion between the EMG and vision evidences.}
\label{graphicV1}
\end{figure}

% The fusion problem can be formulated as:

% \begin{multline}
%     P(L_1 = l_1, L_2 = l_2 \mid E=e, V=v) \approx \\
%     P(L_2=l_2 \mid L_1=l_1, V=v, E=e) \\\times  P(L_1=l_1 \mid E=e) 
%     \times \tfrac{P(L_2=l_2, E=e \mid L_1=l_1, V=v)}{P(E=e \mid L_1=l_1, V=v)} \\
%     = \tfrac{P(E=e \mid L_2=l_2, L_1=l_1, V=v) \times P(L_2=l_2 \mid V=v, L_1=l_1)}{P(E=e \mid L_1=l_1, V=v)}\\
%     = \tfrac{P(E=e | L_2=l_2, L_1=l_1, V=v) \times P(L_2=l_2 \mid V=v) P(L_1=l_1 \mid E=e)}{P(E=e \mid L_1=l_1, V=v)}
% \end{multline}

% Where the final decision is:

% \begin{multline}
%     \{\hat{l_1^{max}}, \hat{l_2^{max}}\} = 
%     \argmax_{l_1, l_2} \{
%     \frac{P(E=e \mid L_2=l_2, L_1=l_1)}{P(E=e \mid L_1=l_1)} \\
%     \times P(L_2=l_2 \mid V=v) \times P(L_1=l_1 \mid E=e) \} \\
%     = \argmax_{l_1, l_2} \{ P(L_2=l_2 \mid V=v) \times P(L_1=l_1 \mid E=e) \}
% \end{multline}

% Where, $l_1$ is the estimated grasp type.

As shown in \autoref{graphicV1}, given the visual information and appearance $V$ of a specific target object, the user first reacts accordingly to the observed $V$ with a designated gesture intent $L$, and then corresponding muscle activities $M$ of the user are triggered and executed according to the intended gesture comprehended by the user. The purpose of the multimodal fusion between EMG and vision was to maximize the probability of the intended gesture given the collected EMG and vision evidences. Therefore, the optimization of this fusion was formulated by the maximum likelihood problem of object $P(L=\hat{l}|V,M)$, modeled by the graphic model in \autoref{graphicV1}, where $V$ and $M$ are defined as vision evidence and muscle EMG evidence, and $L$ presents the grasp type with optimal decision $\hat{l}$.

For deriving the optimization object of the multimodal fusion, we wrote down the joint distribution of $L$, $V$ and $M$ according to the graphic model in \autoref{graphicV1} as follows:
\begin{equation}
\label{eq:jontDistr}
\begin{split}
          P(L,V,M) = P(M|L) P(L|V) P(V),
\end{split}
\end{equation}
so the object $P(L=\hat{l}|V,M)$ of the optimization problem can be further written as Eq.~\eqref{fusion_object} according to Eq.~\eqref{eq:jontDistr}:
\begin{equation}
\label{fusion_object}
\begin{split}
            & \max_{\hat{l} \in \{1,\cdots,13\}} P(L=\hat{l}|V,M)  \\
          = & \max_{\hat{l} \in \{1,\cdots,13\}} \frac{P(M|L=\hat{l}) P(L=\hat{l}|V) P(V)}{P(V,M)}.
\end{split}
\end{equation}

Since $P(V)$ and $P(V,M)$ are not functions of variable $L$ and $P(L)$ is evenly distributed over all classes, the optimization object Eq.~\eqref{fusion_object} is equivalent to the following representation 
\begin{equation}
\label{fusion_object_final}
\begin{split}
          &\max_{\hat{l} \in \{1,\cdots,13\}} P(M|L=\hat{l}) P(L=\hat{l}|V) \\
          = &\max_{\hat{l} \in \{1,\cdots,13\}} \frac{P(L=\hat{l}|M) P(M)}{P(L=\hat{l})} P(L=\hat{l}|V) \\
           \sim & \max_{\hat{l} \in \{1,\cdots,13\}} P(L=\hat{l}|M) P(L=\hat{l}|V).
\end{split}
\end{equation}

The final object of the multimodal fusion is illustrated in Eq.~\eqref{fusion_object_final}, where the optimal estimation $\hat{l}$ of the ground truth should lead to a maximum value of $P(L=\hat{l}|M) P(L=\hat{l}|V)$ among all the $13$ grasp types $l \in \{1,\cdots,13\}$. The probability estimators of $P(L=l|M)$ and $P(L=l|V)$ are implemented by the EMG classifier and CNN built in Section~\ref{sec:emg} and \ref{sec:vision}, respectively.

\section{Experimental Results}
\label{sec:results}

To demonstrate the efficacy of our proposed method, we utilize the dataset collected in \autoref{subsection:data_collection} and train the proposed EMG and visual gesture classifiers, fuse the two using the propose Bayesian evidence fusion and demonstrate the results. We first provide the metrics used for each module in \autoref{subsection:metrics}. Then we present the results of EMG gesture classification utilizing EMG modality in \autoref{subsection:emg_res}, followed by the results from the visual modality in \autoref{subsection:vis_res}. Lastly, results from fusing both modalities are provided in \autoref{subsection:fus_res}. The results provided are all analyzed offline.

\subsection{Metrics}
\label{subsection:metrics}
In assessing our multimodal system, we implement intuitive metrics to gauge the effectiveness of the individual modules and the integrated framework.

For the visual module, the mean Average Precision (mAP) is the metric of choice. The mAP is a comprehensive measure that evaluates the average precision across different classes and Intersection over Union (IoU) thresholds. It reflects the model's accuracy in identifying and classifying various objects in images. By averaging out the precision scores over a variety of classes and IoU benchmarks, mAP provides an overall picture of the model’s performance in detecting objects accurately and consistently. To elaborate more, for each class, predictions are sorted by the confidence score and Intersection over Union (IoU) is calculated. IoUs of over the threshold are considered True Positives (TP) and the rest are False Positives (FP). Then precision (P) and recall (R) are calculated given TP and FP values. For each class, the area under the curve of the P-R curve provides the Average Precision (AP) for that class. Finally, mAP is provided by calculating the mean across AP values of all classes.

In contrast, for the electromyography (EMG) and the fusion modules, we apply the top-1 accuracy metric. Top-1 accuracy is a widely recognized metric in model evaluation, demanding that the model's most probable prediction matches the expected result to be considered correct. This metric is exacting, as it counts only the highest-probability prediction and requires a perfect match with the actual label for the prediction to be deemed accurate. This exactness is crucial for the EMG signals and the precision necessary in the fusion of EMG and visual data for effective robotic control, making top-1 accuracy an appropriate measure of the system’s reliability.

By leveraging these metrics, we ensure a thorough and detailed evaluation of our system's accuracy and reliability. mAP allows us to understand the visual module's complexity in image processing, whereas top-1 accuracy offers a straightforward evaluation of the EMG and fusion modules' operational success. Together, these metrics provide a foundation for ongoing improvements and guide the advancement of our system towards enhanced real-world utility.

\subsection{Dynamic-EMG Classification}
\label{subsection:emg_res}
\subsubsection{Training and Validation}
We performed inter-subject training and validation for the $14$-class gesture classification of dynamic EMG. The classification analyse was implemented through a left-out validation protocol. For each subject, the collected $6$ EMG trials of each object in \autoref{fig:system_b} were randomly divided into training set ($4$ trials) and validation set ($2$ trials), leading to $216$ training trials ($66.7\%$) and $108$ validation trials ($33.3\%$) in total for each subject. The classifier was only trained on the training set while validated on the validation set which was unseen to the model. We validated the pre-trained model on each entire EMG trial (including four phases of reaching, grasping, returning and resting) from the validation set, whereas we trained the model only with reaching, grasping and resting phases in each EMG trial of the training set. Since our main goal is to decode the grasping intention and pre-shape the robotic hand at an earlier stage of reach-to-grasp motion before the final grasp accomplished, we therefore excluded the EMG data of returning phase during training to reduce the distraction of the model from the phase where the hand already released the object.
% All training and validation of the EMG classifier were performed with Python3 using Scikit-learn library.

\subsubsection{Time-Series Alignment}

\begin{figure}[t]
  \centering
  \includegraphics[width=0.4\textwidth]{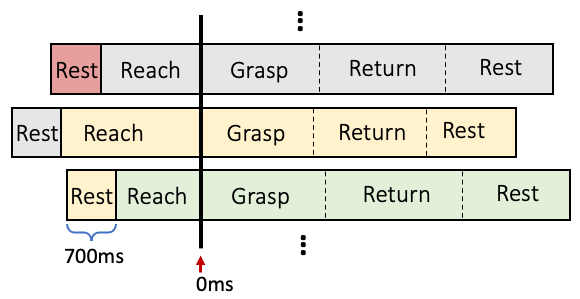}
  \caption{The time series alignment for different trials.}
  \vspace{-0.2cm}
\label{fig_alignment}
\end{figure}

As shown in \autoref{fig_alignment}, time series of all validation trials were aligned with the beginning of the grasping phase, which was marked as $0$ms in the timeline. The overall evaluated performance of the model was averaged over the performances of all validation trials based on the given aligned timeline. The four dynamic phases were freely performed by the subject, leading to their different lengths. Therefore, aligning the validation time series with the grasping phase during the performance average concentrated the assessment more on the central region between reaching and grasping phases, which were the most important phases for decision making. This resulted in more orderly time series, which were more relevant to the dynamic validation of overall performance .

As illustrated in \autoref{fig_alignment}, each validation trial was shifted backward $700$ms (around half the length of entire resting phase), for presenting the resting phase from last trial in front of the reaching phase of current trial, to show the dynamic performance transition between the two movement phases.

\subsubsection{Results}

\setcounter{subfigure}{0}
\begin{subfigure}
\setcounter{subfigure}{0}
    \centering
    \begin{minipage}[b]{0.47\textwidth}
       \includegraphics[width=\textwidth]{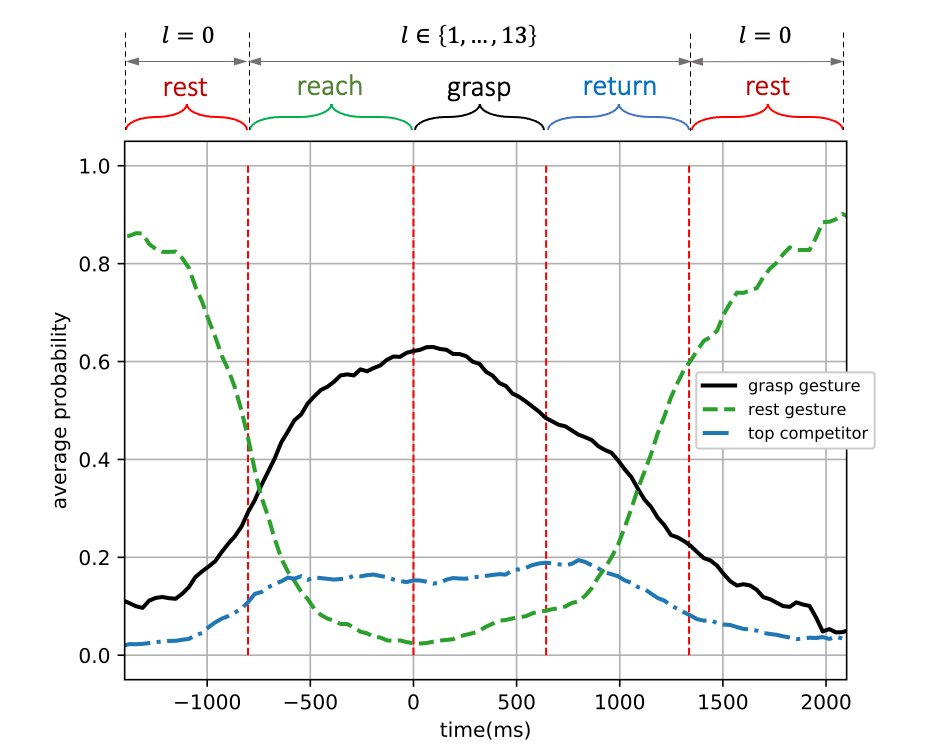}
       \caption{Average validation probabilities}
       \label{fig:EMG_results_a}
    \end{minipage}%
\setcounter{subfigure}{1}
\begin{minipage}[b]{0.45\textwidth}
   \includegraphics[width=\textwidth]{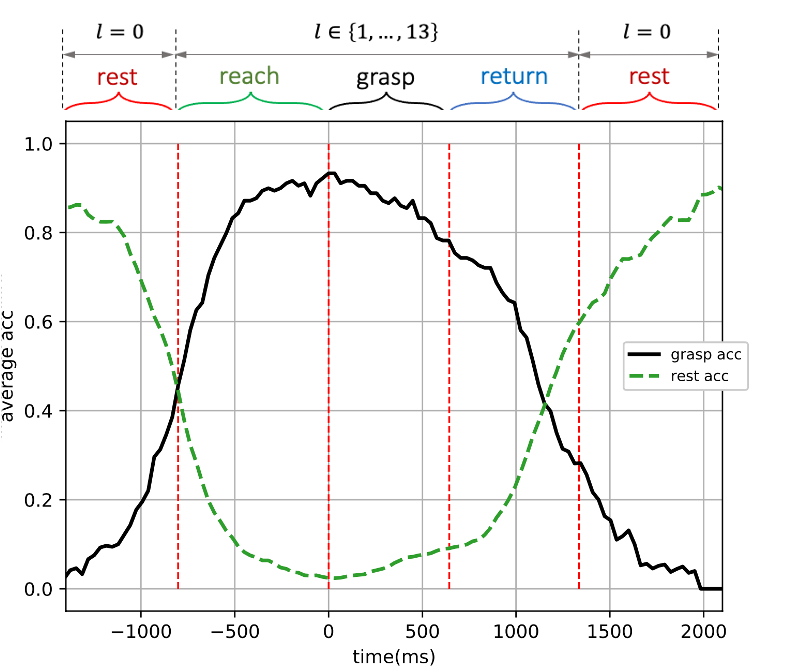}
   \caption{Average validation accuracy}
   \label{fig:EMG_results_b}
\end{minipage}
\setcounter{subfigure}{-1}
\caption{The performance on the validation set of the dynamic-EMG gesture classifier. The grasp gesture and accuracy is defined based on the executed true gesture during the non-resting phases, the rest gesture and accuracy represent the open-palm/rest gesture during the resting phase. The top competitor is the second most likely prediction from the classifier (second best decision the model predicts)}
\label{fig:EMG_results}
\end{subfigure}

Performance of the dynamic-EMG classifier was evaluated by two metrics - the predicted probabilities and the accuracy on validation set, which are presented in \autoref{fig:EMG_results} as functions of time. Each time point in \autoref{fig:EMG_results} represents a EMG window and both metrics were averaged based on each time window over the validation set. The breakpoints between different motion phases (represented by vertical dashed lines) were also averaged across validation trials. In \autoref{fig:EMG_results_a}, the predicted probability is defined as the output probability of the classifier corresponding to each class, and here we show the probabilities of grasp gesture, rest gesture and top competitor. The grasp gesture is defined as the executed true gesture during the non-resting phases, the rest gesture represents the open-palm/rest gesture, and the top competitor is identified as the most possible gesture except for the grasp gesture and the rest gesture. In \autoref{fig:EMG_results_b}, the corresponding accuracy curves of successfully detecting the grasp gesture and rest gesture are displayed, where the accuracy is defined as the frequency of appearance of a specific label with the maximum probability over the output probability distribution.

% The performance of the dynamic-EMG classifier is shown in \autoref{fig:EMG_results} as function of time, to inspect the performance variation during different dynamic phases. In \autoref{fig:EMG_results}, the predicted probabilities and accuracy on validation set of the dynamic-EMG gesture classifier are respectively presented. The grasp gesture is defined as the executed true gesture during the non-resting phases, the rest gesture represents the open-palm/rest gesture during the resting phase, and the top competitor is identified the most possible gesture except for the executed true gesture and the rest gesture. The predicted probabilities of the grasp gesture, rest gesture and top competitor on validation set for each time point within a trial are shown in \autoref{fig:EMG_results} (a), while in \autoref{fig:EMG_results} (b) the corresponding accuracy curves of successfully detecting the grasp gesture and rest gesture are displayed. Each time point in \autoref{fig:EMG_results} (a) and (b) represents a EMG window and the performance was averaged within the same window over all validation trials. The displayed breakpoints between different motion phases (represented by vertical dashed lines) were also averaged across all validation trials.

As illustrated in \autoref{fig:EMG_results_a}, the predicted probability of the grasp gesture $l \in {\{0,1,...,13\}}$ increased steadily during the reach-to-grasp movement when the grasp was carried out from the resting status, reaching its peak in the grasping phase, and then gradually decreased when subject finished the grasp and returned to resting status again. Simultaneously, the predicted probability of rest gesture first reduced dramatically to the value lower than $0.2$ as the grasp movement happened, until the hand returned to the resting position when the open-palm probability progressively went up again. In addition, the predicted probability of the top competitor was remained stably lower than $0.2$. 
% The top competitor was defined as the gesture with highest predicted probability among all the other $12$ gestures excluding the executed gesture and rest gesture for each EMG window. As demonstrated in \autoref{fig:EMG_results} (a), the predicted probability of the top competitor was remained stably lower than $0.2$. 
The first intersection of predicted grasp-gesture and rest-gesture probability curves indicates the point when the grasp-gesture decision outperforms the rest-gesture decision. 
Ideally, this intersection is expected to appear right at the junction where the resting phase ends and the reaching phase starts in order to indicate the beginning of the hand motion. 
However, in practice, the hand movement could only be predicted based on the past motion, and such motion first starts from the reaching phase. So the intersection is expected to be observed after the start point of the reaching phase, but the closer to this start point, the better. In \autoref{fig:EMG_results_a}, the first intersection of the two curves appears $>700$ms earlier than the start of the grasping phase, which is after but very close to the beginning of the reaching phase and allows enough time to pre-shape the robotic hand before the actual grasp. 

As a validation metric, we use the top-1 accuracy as the conventional accuracy, where model prediction (the one with the highest probability) must be exactly the expected answer. As shown in \autoref{fig:EMG_results_b}, for the grasp gesture classification, the averaged accuracy was higher than $80\%$ throughout most of the reaching and grasping phases, which are the most critical phases for making robotic-grasp decision. The average accuracy during resting phase were also highly accurate and sensitive to perform as a detector to trigger the robotic grasp as shown in \autoref{fig:EMG_results_b}. In between dynamic phases of resting and non-resting, the accuracy also shows a smooth transition. It is worth noting that the validation accuracy was still higher than $75\%$ at the beginning of the returning phase even though the model was not trained on any data from that phase, illustrating the generalization and robustness of our model on dynamic EMG classification.

\subsubsection{Inter-Subject Analysis}
\begin{table}[t]
\centering
\begin{tabular}{@{}lc|ccc|c@{}}
\toprule
        &             & \textbf{Fold 1} & \textbf{Fold 2} & \textbf{Fold 3} & \textbf{Mean} \\ \midrule
\multirow{2}{*}{Sbj. 1} & motion clf. & 77.2\%          & 74.5\%          & 78.2\%          & 76.6\%        \\
        & gesture clf. & 91.5\%          & 88.7\%          & 91.5\%          & 90.6\%        \\ \midrule
\multirow{2}{*}{Sbj. 2} & motion clf. & 75.3\%          & 73.1\%          & 72.8\%          & 73.7\%        \\
        & gesture clf. & 88.9\%          & 85.3\%          & 86.6\%          & 86.9\%        \\ \midrule
\multirow{2}{*}{Sbj. 3} & motion clf. & 72.6\%          & 73.4\%          & 72.7\%          & 72.9\%        \\
        & gesture clf. & 87.4\%          & 87.2\%          & 87.8\%          & 87.5\%        \\ \midrule
\multirow{2}{*}{Sbj. 4} & motion clf. & 78.7\%          & 76.6\%          & 76.4\%          & 77.2\%        \\
        & gesture clf. & 89.3\%          & 87.4\%          & 88.3\%          & 88.4\%        \\ \midrule
\multirow{2}{*}{Sbj. 5} & motion clf. & 78.4\%          & 75.9\%          & 79.1\%          & 77.8\%        \\
        & gesture clf. & 82.8\%          & 80.5\%          & 82.2\%          & 81.8\%        \\ \bottomrule
\end{tabular}
\caption{Inter-subject motion and gesture classification accuracies among the 5 subjects.}
\label{tab:intersub}
\end{table}

To analyze the inter-subject variations in each subject, results in \autoref{tab:intersub} demonstrate the average classification accuracies for each cross-
validation fold from both motion-phase classifier and gesture classifier, with the mean accuracy over
all folds given for each subject. The average accuracy of the 4-class motion-phase classifier for each individual subject varies between 72.9\% and 77.8\%, while the 14-class gesture classifier presents
average accuracies ranging from 81.8\% to 90.6\%. Those results reveal that the grasp phases and
types were well-predicted in general. The dynamic-EMG gesture identification showed a better performance than the motion-phase detection, due to the higher degree of freedom regarding to how subjects performed different motion phases than the grasp type, as the experiment protocol did not specify the particular speed or angle to grasp. The illustrated inter subject variability across different validation trials may also come from the varied grasping patterns and directions for different trials of the same subject. However, this higher degree of freedom 
could enable more robustness and stability of the model to a wider range of user postures during the dynamic grasp activity. In addition, the training data from various subjects also present different classification performances, which could be influenced by factors such as shifting sensor locations and distinct movement patterns of different users.

\subsection{Visual Gesture Classification}
\label{subsection:vis_res}
We fine-tune the pre-trained YoloV4 on our dataset using the images augmented by our copy-paste. These images are further generalized by utilizing photometric and geometric distortions and other augmentations from "bag of freebies" \citep{yolov4} i.e., saturation, exposure, hue and mosaicing. To make the network completely invariant to the object's color, we set hue to the maximum value of 1.0 in our experiments. All the training setup are outlined at \autoref{tab:train_setup}.

\begin{table}[!t]
\renewcommand{\arraystretch}{1.3}
\caption{Training Setup}
\label{tab:train_setup}
\centering
\begin{tabular}{|>{\bf}c||c|}
    \hline
    Learning Rate & $0.001$\\
    \hline
    Momentum & $0.949$\\
    \hline
    Decay & $0.0005$\\
    \hline
    Iterations & $28680$\\
    \hline
    Batch Size & $64$\\
    \hline
    Input Size & $608\times608\times3$ \\
    \hline
    Angle & $0$\\
    \hline
    Saturation & $5$\\
    \hline
    Exposure & $1.5$\\
    \hline
    Hue & $1.0$\\
    \hline
\end{tabular}
\end{table}

\begin{figure}[t]
  \centering
  \includegraphics[width=0.44\textwidth]{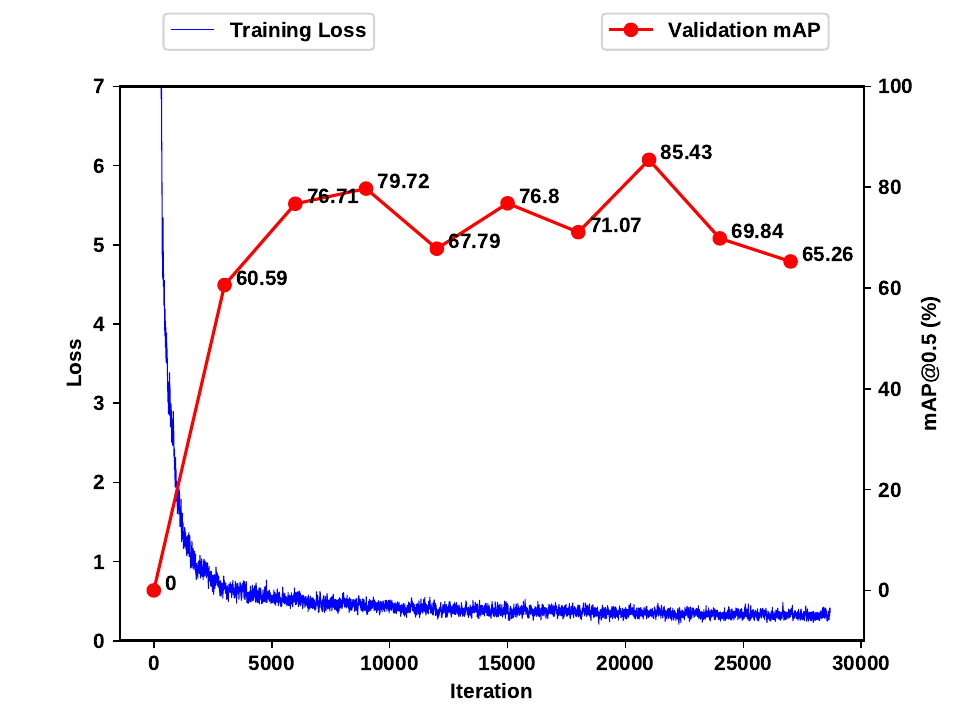}
  \caption{Training and validation loss (mAP). As the computation time for each mAP calculation will significantly impact the training time of the network, mAP calculations happen every 3000 iterations.}
  \vspace{-0.2cm}
\label{fig:yolo_chart}
\end{figure}

To train, validate and test the network, imagery data has been split based on 4 trials for training and validation and 2 trials for test, to 48537 ($54.1\%$), 12134 ($13.5\%$) and 29029 ($32.3\%$) images respectively, and balanced according to each class. The augmented data is only used in the training process and is applied in an in-place style. This means that the data augmentation is applied directly to the existing data instances without creating additional copies. This can be particularly useful when computational resources or storage capacity is a concern, as it allows for the expansion of the dataset without significantly increasing its size. The transformations are applied on-the-fly during the training process, and each epoch of training sees a slightly different version of the data which enables improved model generalization by presenting a more varied dataset, helping the model perform better on unseen data. We use the commonly used mAP for the visual module where mAP calculates the average precision across different classes and/or Intersection over Union (IoU) thresholds, providing a single comprehensive measure of a model's performance in detecting and classifying various objects in images. Training loss and validation mean average precision (mAP) are outlined in \autoref{fig:yolo_chart}. To prevent any over-fitting, validation mAP provides a guide on the iteration with the best generalization, reaching $85.43\%$ validation mAP. This results in the very close mAP of $84.97\%$ for the test set, proving the high generalization of the network. Notably, this marks a substantial improvement compared to the baseline, with the original COCO dataset yielding a 64.4\% mAP. The close alignment of our results with the COCO baseline underscores the success of transfer learning, emphasizing the data's intrinsic similarity to COCO.

\subsection{Multimodal Fusion of EMG and Vision}
\label{subsection:fus_res}

To have a fair comparison of accuracy between EMG and visual classifiers and their resulting fusion, each classifier is trained on the same set of data and tested on data that is unseen to all classifiers. To this end, from the 6 trials belonging to each experiment, 4 has been randomly selected for the training of EMG and visual classifiers and the remaining 2 as the test data. As a result, all of the results presented in this work are based on data unseen to both EMG and visual classifiers.

\begin{figure}[t]
  \centering
  \includegraphics[width=0.49\textwidth]{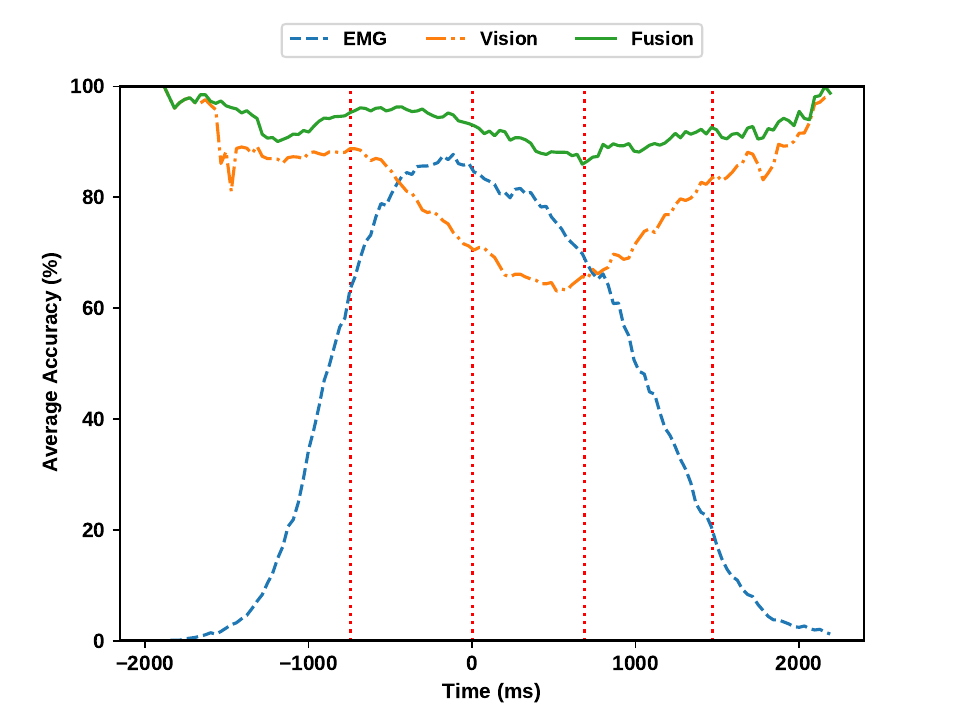}
  \caption{Average validation accuracy. Note that the predictions from each source generally complements the other source. Fusing EMG and visual evidence has improved the overall accuracy and robustness of the estimation.}
\label{fig:fusion_acc}
\end{figure}

\autoref{fig:fusion_acc} visualizes the average validation accuracy of EMG, vision and fusion modules over time. The accuracy at each time is defined as the frequency of appearance of the correct label as the maximum probability in a classifier's output probability distribution. We observe that the classification of visual information can perform decently almost at all times and without significant changes except during the grasp phase and some portions of the neighbouring phases where the object is most likely occluded. On the other hand, EMG information can complement this deficiency, given that the subject's hand is mostly active during this phase. This is clearly evident in \autoref{fig:fusion_acc} as the EMG classification outperform visual classifier. The complementary characteristic of EMG and visual information is also noticeable at rest phases where the subject's hand is least active. During resting, the object of interest is clearly visible by camera, therefore resulting in high accuracy of the visual classifier.

In addition to this complementary behaviour, fusion is always outperforming each individual classifier. This means that fusion can add additional robustness even when both sources provide enough information for a correct decision. To provide more details, the summary of each module's accuracy is provided in \autoref{tab:acc_per_phase} in different phases. As outlined in the aforementioned table, the grasp classification accuracy while solely utilizing the Gesture classifier on the EMG modality is 81.64\% during the reaching phase. On the other hand, solely relying on the visual modality, the visual classifier yields 80.5\% accuracy during the reaching phase. Combining these two provides a significant improvement of 95.3\% accuracy which results in 13.66\% improvement for the EMG modality and 14.8\% improvement compared with the visual modality alone.

Having a robust control of the grasp type at all times is essential especially at reaching phase, where the actual grasping decision is sent to the robot's actuators. Fusion of visual and EMG evidence enables robust classification of grasp types, giving the robotic hand enough time to perform the grasp. The existing fusion method operates exclusively based on the instantaneous outcomes of each modality, rendering it stationary in time and, in essence, memory-less. This approach entails making decisions at a specific moment without considering the historical context of past decisions. To introduce temporal dynamics and incorporate the influence of prior decisions, one could explore methodologies such as a Kalman filter or a neural network. These mechanisms have the capacity to leverage the history of past decisions, allowing the fusion process to be more adaptive and informed by the temporal evolution of the data. Therefore, we recommend utilizing robot control policy to exploit past decisions into their fusion based on their system configurations and constraints, as our experiments show that by simply smoothing fusion decisions, the average accuracy is further increased to the significant value of $96.8\%$. We suggest that future studies can also utilize more sophisticated methods based on machine learning and deep learners for fusion of the information.

\begin{table}[t]
\centering
\renewcommand{\arraystretch}{1.3}
\begin{tabular}{c|c|c|c|c|c|c|}
\cline{2-7}
 & \multicolumn{5}{c|}{\textbf{Phase}} & \multirow{2}{*}{\textbf{Total}} \\ \cline{2-6}
 & \textbf{Rest} & \textbf{Reach} & \textbf{Grasp} & \textbf{Return} & \textbf{Rest} &  \\ \hline
\multicolumn{1}{|c|}{\textbf{EMG}} & 16.86 & \textbf{81.64} & 78.66 & 45.41 & 6.33 & 41.85 \\ \hline
\multicolumn{1}{|c|}{\textbf{Vision}} & 90.69 & \textbf{80.5} & 66.22 & 74.05 & 88.59 & 81.46 \\ \hline
\multicolumn{1}{|c|}{\textbf{Fusion}} & 94.89 & \textbf{95.3} & 89.7 & 89.81 & 93.31 & 92.93 \\ \hline
\end{tabular}
\caption{Accuracy of each module during different phases in percentage. Reach phase is demonstrated in bold, as is the most critical phase for decision making. Note that a random classifier has a $\frac{1}{13}=7.7\%$ chance for each class.} 
\label{tab:acc_per_phase}
\end{table}

\section{Discussion}
\label{sec:discussion}

Robotic prosthetic hands hold significant promise, particularly considering that limb loss often occurs during working age. Dissatisfaction with the effectiveness of a prescribed prosthesis can adversely impact an amputee's personal and professional life. Consequently, a functional prosthesis is vital to mitigate these challenges and enhance the quality of life for amputees. However, single-sensing systems in prosthetics come with inherent limitations. In the utilization of robotic prosthetic hands for transradial amputees, intuitive and robust control system that can compensate for the challenges posed by missing or inaccurate sensor data is paramount. This research advocates for a shift from relying on single data sources, such as EMG or vision, which each have their own limitations, to a multimodal approach that fuses various types of information. 

In controlling robotic prosthetic hands, the intended grasp type needs to be known a few hundred milliseconds before the grasp phase. Therefore, in addition to inferring the grasp type desired by the user, the time when this information is obtained is important. Hence, our article investigates and analyzes the evidence available for inferring grasp type over time. To aid with the time understanding, e.g., to enable the robotic hand to actuate the fingers at the correct time, which is crucial for successful robotic grasping, our work employs an EMG phase detection algorithm in addition to the EMG and vision grasp classifiers. The focus on real-time understanding and analysis is an important aspect that sets this work apart from the current state-of-the-art.

The classification accuracies for EMG, vision, and their fusion across different phases provides insightful results. The EMG classification shows varied performance, excelling in phases like Reach (81.64\%) and Grasp (78.66\%) where the EMG data is most meaningful, but significantly lower in Rest (6.33\%) where the data semantically has no correlation with the grasp being performed. Here we observe that the low accuracy attained in the Rest phase by EMG is aligned with a random classifier where the chance of a random classifier for correctly classifying a class among 14 labels is $\frac{1}{14}=7\%$. This reiterates that no useful information can be utilized from the EMG modality during the rest phase and our EMG classifier is not overfitted to the noise. Vision-based classification, on the other hand, maintained high accuracy across most phases, particularly in Rest (90.69\%) where the object has no occlusions.

Most notably, the fusion of EMG and vision data consistently achieved superior accuracy in all phases. Considering the reaching phase as the most accurate case, while the EMG modality provides 81.64\% and the visual modality yielding 80.5\% accuracy when used individually, combining these two provides a significant improvement of 95.3\% accuracy which results in 13.66\% improvement for the EMG modality and 14.8\% improvement compared with the visual modality alone. This highlights the significant advantage of combining these modalities, particularly in critical phases like Reach and Grasp, where precise control is of supreme importance. The fusion method's robustness across different phases, with accuracies consistently above 89\%, underscores its potential in enhancing the functionality and reliability of prosthetic hand control.

In our work we have introduced several novel advancements in grasp classification that are not observed in prior research. These enhancements are particularly evident in our approach to visual generalization, finer distinction of classes, and provision of critical timing information. Firstly, in terms of visual generalization, we utilize background generalization to provide more realistic data for our visual grasp classification module. This step, not previously observed in grasp classification research, allows our system to better prepare for real-world scenarios, making our data more representative and robust.

While direct comparisons to state-of-the-art are challenging due to differing data, our work further distinguishes itself through a more granular classification of grasp types. We identify 14 distinct grasp types, compared to the 10 grasp types utilized in \citep{cognolato2022improving}. This expanded classification presents a more challenging and realistic problem, advancing beyond the current state-of-the-art. Our method not only provides a more refined solution but also achieves a higher accuracy in the critical reaching phase. We report a 95.6\% accuracy during the reaching phase, compared to the 70.92\% average accuracy for able-bodied subjects reported in \citep{cognolato2022improving}.

Lastly, and most importantly, our research significantly extends the scope beyond the conventional focus on rest and non-rest phases, as seen in previous studies. Our study investigates the time dynamics and the different phases of electromyography (EMG) signals within our setup to enhance the classification of grasp types. It is crucial to understand how system components react during different phases of EMG; the way the hand is resting has no correlation to how the hand will grasp later. Hence, resting should not be part of the grasp classifier but a separate model. Neglecting this consideration may lead to overfitting of the model, which can negatively impact its generalizability. Similar limitations have been observed previously with spatial domain, where models trained solely on cloudy images tended to incorrectly identify tanks \citep{tanks}. 

Our work not only explores the implementation of grasp intent inference but also lays the groundwork for practical robotic implementation. This is achieved by carefully analyzing the various phases of object interaction, including precise estimations of when to initiate and cease interaction. Our approach enables the detection of all four critical phases involved in object handling: reach, grasp, return, and rest. This comprehensive analysis is evident throughout our research, influencing our protocol design, the selection of inference data, and the development of our inference and fusion methodology. The depth of our approach is particularly apparent in our discussions on accuracy over time. Therefore, this work provides a new avenue in grasp classification in the field providing more precise and meaningful grasp classification. 

Our advancements represent a significant leap forward in grasp classification, particularly in terms of accuracy and realism in the critical phases of prosthetic hand operation.

\subsection*{Limitations and Future Advancements}

While we recognize that a larger sample size would contribute to enhanced generalizability, we would like to emphasize that our study serves as a preliminary investigation into the feasibility and efficacy of the proposed multimodal fusion approach. The decision to begin with a smaller cohort was based on the exploratory nature of our work but overall would not make a large impact on the described system for data fusion which is the central focus of the paper. This approach allowed us to assess the initial viability of our methodology and pave the way for more extensive studies in the future.

The current study has focused on establishing the feasibility and efficacy of our approach in a controlled environment with healthy subjects. However, we understand the importance of addressing the translational aspect to clinical applications, particularly in the context of amputees. To shed light on the translatability of our approach, we acknowledge that training our model for amputees would necessitate a subject-specific adaptation. Each amputee presents a unique set of physiological characteristics and muscle activation patterns, requiring a tailored training process for optimal performance. Future work in this direction would involve subject-specific training.

It is noteworthy to mention that measurement of EMG from intrinsic hand muscles used in this experiment (FDI, APB, FDM, and EI) would likely not be present in most amputees requiring a prosthetic hand. The remaining muscles used (EDC, FDS, BRD, ECR, ECU, FCU, BIC, and TRI) would remain in most wrist level amputees. In the case of most below elbow amputees the residual limb retains key anatomical landmarks, allowing for strategic placement of electrodes. Several investigations have compared classification performance using targeted vs non-targeted electrode placement with outcomes generally favoring targeted placement \citep{yoo2019myoelectric, farrell2008comparison}. Furthermore, most commercial myoelectric devices offer some ability for customization of electrode position though typically for a small number of electrodes. Additionally, numerous investigation have examined the impact that number of electrodes has on classification and there are currently several approaches for determining pareto-optimality between electrode number and classification accuracy. Electrode placement and number were not specific foci of the current investigation. In general electrode placement will be determined by the anatomy of the amputee and other clinical considerations hence beyond the scope of this work. Certainly, as the methods proposed here progresses to real application the number and placement of electrodes could be considered. However, this does not diminish our current contribution because the system to implement fusion is not dependent on these factors.

Lastly, there are several recent works incorporating palm-mounted cameras not only to recognize the object being grasped, but also the reaching conditions e.g., the wrist orientation thereby facilitating a more versatile grasp with additional degrees of freedom \citep{cirelli2023semiautonomous, castro2022hybrid}. While head-mounted cameras offer a wider, more natural field of view that is generally more stable and reliable for grasp detection, it is beneficial to employ an additional palm-mounted camera enabling the recognition of optimal approaching conditions directly from the hand's viewpoint and consequently contributing to a more autonomous operation of the robotic hand.

\section{Conclusion}
\label{sec:conclusion}

For robotic prosthetic hands to effectively compensate for the lost ability of transradial amputees during daily life activities, control of the hand must be intuitive and robust to missing and sometimes inaccurate sensor data. Solely relying on one source of information e.g., EMG or vision, is prone to poor performance due specific drawbacks of each source. Hence a shift in the approach to one that fuses multiple sources of information is required. In this work we collected a dataset of synchronized EMG and visual data of daily objects and provided details on our proposed process for sensor fusion including EMG segmentation and gesture classification and camera-based grasp detection that is bundled with background generalization using copy-paste augmentation. Based on a graphical model, we represented the multimodal fusion as a maximum likelihood problem to increase robotic control accuracy and robustness.

In our experiments, we observed the complementary behaviour of visual and EMG data. EMG generally performs better when reaching and grasping an object when the imagery data cannot provide useful information due to occlusion. Visual information can provide information about the needed grasp prior to movement, when EMG is unavailable. Our experiments show that fusion always outperforms each individual classifier demonstrating that fusion can add additional robustness even when both sources provide enough information for a grasp decision. We observe that fusion improves the average grasp classification accuracy while at reaching phase by 13.66\%, and 14.8\% for EMG (81.64\% non-fused) and visual classification (80.5\% non-fused) respectively to the total accuracy of 95.3\%.

\section*{Acknowledgment}
This work is partially supported by NSF (CPS-1544895, CPS-1544636, CPS-1544815). 

% \typeout{}
\bibliographystyle{Frontiers-Harvard}
\bibliography{refs}

\end{document}